\documentclass{article}
\usepackage{ijcai23}

\usepackage{times}
\usepackage{soul}
\usepackage{url}
\usepackage[hidelinks]{hyperref}
\usepackage[utf8]{inputenc}
\usepackage[small]{caption}
\usepackage{graphicx}
\usepackage{amsmath}
\usepackage{amsthm}
\usepackage{booktabs}
\usepackage{subfig}
\usepackage{mathbbol}
\usepackage{booktabs,multirow}
\usepackage{subcaption}
\usepackage{caption}
\usepackage[ruled,linesnumbered]{algorithm2e}
\usepackage[switch]{lineno}
\usepackage{svg}

\title{\textit{FedFa}: A Fully Asynchronous Training Paradigm for Federated Learning}

\author{
Haotian Xu$^1$
\and
Zhaorui Zhang$^1$ *\and
Sheng Di$^2$\and
Benben Liu$^3$\and \\
Khalid Ayed Alharthi$^4$\And
Jiannong Cao$^1$, \textit{IEEE Fellow}
\affiliations
$^1$The Hong Kong Polytechnic University, Hong Kong\\
$^2$Argonne National Laboratory, USA\\
$^3$The University of Hong Kong, Hong Kong\\
$^4$University of Bisha, Saudi Arabia
\emails
haotian.xu@connect.polyu.hk, \{zhaorui.zhang, jiannong.cao\}@polyu.edu.hk, \\
sdi1@anl.gov, benbenliu@hku.hk,
kharthi@ub.edu.sa
}
\begin{document}

\maketitle

\begin{abstract}
Federated learning has been identified as an efficient decentralized training paradigm for scaling the machine learning model training on a large number of devices while guaranteeing the data privacy of the trainers. FedAvg has become a foundational parameter update strategy for federated learning, which has been promising to eliminate the effect of the heterogeneous data across clients and guarantee convergence. However, the synchronization parameter update barriers for each communication round during the training significant time on waiting, slowing down the training procedure. Therefore, recent state-of-the-art solutions propose using semi-asynchronous approaches to mitigate the waiting time cost with guaranteed convergence. Nevertheless, emerging semi-asynchronous approaches are unable to eliminate the waiting time completely. 

We propose a full asynchronous training paradigm, called \textit{FedFa}, which can guarantee model convergence and eliminate the waiting time completely for federated learning by using a few buffered results on the server for parameter updating. Further, we provide theoretical proof of convergence rate for our proposed \textit{FedFa}. Extensive experimental results indicate our approach effectively improves the training performance of federated learning by up to $6\times$ and $4\times$ speedup compared to the state-of-the-art synchronous and semi-asynchronous strategies while retaining high accuracy in both IID and Non-IID scenarios.
\end{abstract}

\let\thefootnote\relax\footnotetext{* Zhaorui Zhang is the corresponding author. This paper has been accepted by IJCAI 2024.}

\vspace{-8pt}
\section{Introduction}

Federated learning has achieved great success in recent years thanks to its data privacy protection mechanism while also facing challenges, such as data heterogeneity and communication bottlenecks. The goal of federated learning is to train a good machine learning model to be used by all participants together under the condition that each participant doesn't send out their own data. It not only greatly expands the size of the available training dataset but also utilizes the computational resources of each participant. However, a price often arises from what is gained. Federated learning gains data privacy protection but faces more challenges in communication, resource scheduling, and data heterogeneity\cite{kairouz2021advances}. In federated learning, the training clients are often located in different places connected by low bandwidth networks. The training data set is also diverse across different clients, which affects the convergence behaviors of federated learning significantly. 

% As one emerging distributed learning paradigm, the synchronization component is also a critical part of federated learning, similar to the classical distributed machine learning systems. 
To deploy federated learning in practice, it is necessary for the federated learning systems not only to be accurate but also to satisfy a number of pragmatic constraints regarding issues such as efficiency and system performance. Emerging synchronous parameter update strategies, such as FedAvg, \cite{mcmahan2017communication} and its variants, have become fundamental parameter update strategies in federated learning. However, due to the arithmetic and communication heterogeneity of the devices, the time required to complete the training task and transmit the model parameters to the server varies significantly across clients.
%which is the key point to determine whether a system is appropriate to be deployed in practice. 
Specifically, in FedAvg, the server needs to wait for all the selected clients to complete local training and report their model updates before aggregation. Such a design results in the time required for each round of federated learning training being determined by the slowest client, which leads to great waiting time and inefficiency in computational resources. 

% To reduce the communication overhead, a lot of works have been proposed to compress the gradient and reduce the communication frequency \cite{alistarh2017qsgd,luping2019cmfl,zhang2022mipd,}. 

Scheduling the parameter update in an asynchronous paradigm \cite{zhang2022mipd} is a new direction aiming to reduce the wall-clock time with a well-preserved target accuracy. The asynchronous parameter update strategy eliminates wait times at the system level, where the server does not need to wait for all its selected clients to report their local training results for the aggregation and will update the global model as soon as the model updates from one client arrives \cite{chai2020fedat,zhang2021sapus}. However, the distribution of client speeds is often heavy-tailed in federated learning. Fast clients update the global model frequently, while slower clients make very little progress with the global model. The slower clients often report outdated training results to the server, affecting the global model's performance. To address this issue, most of the recent asynchronous strategies are semi-asynchronous \cite{nguyen2022federated,su2022asynchronous} or semi-synchronous \cite{zhang2022momentum} strategies, where the server will buffer several model updates from clients and then aggregate to update the parameters. These solutions, however, can only reduce part of the waiting time of global barriers but cannot eliminate them.

\textit{Why do the asynchronous federated learning approaches nowadays almost not use a fully asynchronous scheme?} We identify three challenges for the fully asynchronous parameter update strategy in federated learning environments.

\textit{Firstly}, the slower participant clients affect both the synchronous and asynchronous parameter update strategies, which causes long barrier times for model aggregation on the server for the synchronous scheme and results in the model performance degradation for asynchronous mode due to its large staleness. How to migrate the effectiveness of the slower participant clients for the performance of the global model is challenging in the asynchronous parameter update scheme in federated learning.

\textit{Secondly}, federated learning is designed to meet each data owner's requirements for secure transmission and privacy protection, while the privacy protection will be broken when adopting the fully asynchronous parameter update strategies. The client updates the parameters with the server individually, making it easy for other clients to access the updated information from this client in the vanilla fully asynchronous parameter update strategy, which results in information leakage. The semi-asynchronous algorithm eliminates such information leakage by aggregating model updates from several clients, which is the same as synchronous strategies. However, semi-asynchronous and semi-synchronous parameter update strategies bring strong synchronization overhead, thus slowing down the training.

\textit{Thirdly}, providing theoretical proof and guaranteeing the convergence of asynchronous parameter update strategies is challenging. This is caused by the uncertain sequence in which the server updates parameters from clients, which is heavily influenced by the network status and computation resources of participating clients.

In this article, to address the above challenges, we provide new insights about the three-parameter update paradigms for federated learning: synchronous, asynchronous, and semi-asynchronous and design a fully asynchronous parameter update strategy for federated learning, called \textit{FedFa}. 

We summarize our contributions as follows:

\begin{itemize}
\item We propose a fully asynchronous parameter update strategy without any barrier setting for federated learning, called \textit{FedFa}, which updates the global model on the server once it receives an update request from clients, thus eliminating waiting time completely to improve the training performance (different from FedBuff). To mitigate the impact of model updates from slower clients, \textit{FedFa} merges multiple historical model updates into currently received updates through a sliding window.

\item We conduct an in-depth statistical analysis for the convergence rate of our proposed \textit{FedFa}, deriving a theoretical bound, which can be extensively used to guarantee the quality of service in practice.

\item We perform a comprehensive evaluation for \textit{FedFa} on the most popular models, including the Language Model (Bert), CNN model (ResNet18), and Language Model FineTune, in both IID and Non-IID scenarios. Evaluation results show that \textit{FedFa} improves the wall clock time by a factor of $6\times$ and $4\times$, respectively, and the number of communication rounds by a factor of $1.4\times$ and $1.9\times$, when compared to state-of-the-art synchronization and semi-synchronization methods.

\end{itemize}

\section{Background, Related Work and Motivations}

\begin{table}[ht]
%\footnotesize
\scriptsize
    \centering
    \caption{Frequently Used Notations in this Article}
    \captionsetup{font=scriptsize}
    \vspace{-8pt}
\scalebox{0.95}{
\begin{tabular}{cl}
\toprule
Symbol& Description\\
\midrule
$k$& a certain client\\
%\hline
$[m]$&the full set of clients\\
%\hline
$w_g^t$&  $t_{th}$ version parameters on server\\
%\hline
$w_k^{t-\tau_k(t)}$& $t-\tau_k(t)$ version parameters on server\\
%\hline
$\tau_k(t)$&staleness\\
%\hline
$\tau_{max}$&the max staleness\\
%\hline
$S^t$&server's update buffer at phase $t$\\
%\hline
$K$&size of the \textit{Sliding Window}\\
%\hline
$\eta_g$&global learning rate\\
%\hline
$\eta_l^{q}$&local learning rate in step q\\
%\hline
$q$&the $q_{th}$ local steps\\
%\hline
$Q$& the number of local update steps \\
%\hline
$T$& communication rounds for converging \\
%\hline
$\nabla F_k(w)$&gradient calculated on client $k$\\
%\hline
$g_k(w ; \zeta_k))$&stochastic gradient\\
%\hline
$\alpha$& data heterogeneity coefficient\\
%\hline
$\beta_t$&aggregation weight for FedAsync\\
%\hline
$\triangle_k^t$& accumulated gradients at $t_{th}$ steps from client $k$\\
%\hline
$s(\cdot)$& function of staleness for adaptive $\beta_t$\\
%\hline
$f(\cdot)$& loss function \\
%\hline
$\frac{1}{T} \sum_{t=0}^{T-1}\left\|\nabla f\left(w^{t}\right)\right\|^2$& the convergence rate\\
%\hline
$L,\sigma_l,\sigma_g,G$&symbolic upper bounds used in the proof\\
%\hline
$E$& the number of local epochs\\
%\hline
$lb$&local batch size \\
\bottomrule
\end{tabular}
}
\vspace{-12pt}
\label{notations}
\end{table}

\subsection{Synchronous Federated Learning} 
Federated learning uses the synchronous parameter update strategy of when it was first proposed. However, as Federated Learning is put into practical use, the problem of wait times due to device heterogeneity becomes more and more serious. The synchronization paradigm can reduce wait times in several ways. For example, Fedprox \cite{li2020federated} adaptively adjusts the strength of the training tasks assigned to different clients based on their computational power, such as freezing a certain percentage of model parameters based on the computing power of different clients to speed up training. All these methods aim to align the training time of different clients to reduce the waiting time. However, there are some problems with such approaches. On the one hand, such methods either require fine-tuning a very large number of parameters about the device or require the user to elaborate usage schemes, which is not easy to realize in practice. On the other hand, it may lead to additional communication consumption due to the increase in communication rounds. Moreover, this type of method can only reduce, but not eliminate, the waiting time since it is still a synchronized parameter update strategy.

\begin{figure}[ht]
\captionsetup{font=small}
	\centering
	\includegraphics[width=0.8\linewidth, height=0.45\linewidth]{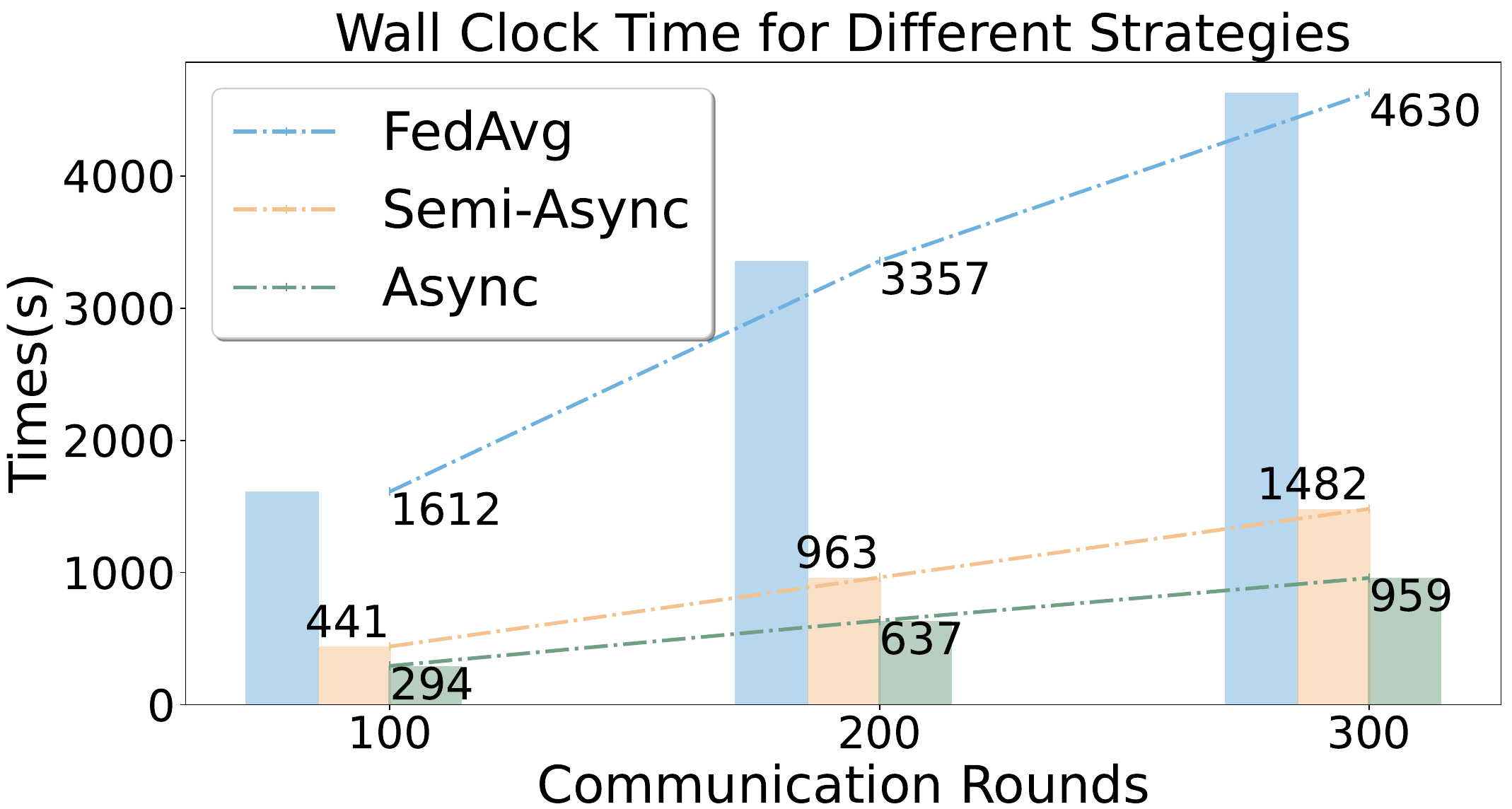}
  \vspace{-6pt}
	\caption{The wall clock time comparison.}
	\label{comm_time}
 \vspace{-8pt}
\end{figure}

\subsection{Semi-asynchronous and Asynchronous FL}

The synchronized parameter update strategy is not necessary for federated learning. Semi-asynchronous and asynchronous algorithms can greatly reduce or even eliminate waiting times right at the system level. Fig.\ref{comm_time} shows the wall clock time required for the three parameter update strategies to complete the same amount of federated learning training. 
FedBuff \cite{nguyen2022federated} is the most popular semi-asynchronous federated learning algorithm nowadays. FedBuff waits for several clients to complete the training task before performing parameter updates, which is \textbf{different} from our \textit{FedFa}.

Fully asynchronous algorithm, Fedasync performs aggregation every time it receives an updated parameter, and the aggregation process is shown in the formula (\ref{fedasync}), where $w_g^{t}$ and $w_l^t$ indicate the parameters on the server and the parameters received by the server from a client at $t_{th}$ steps, respectively. The aggregation weight $\beta_t$ is calculated based on the staleness of the parameters by multiplying the initial aggregation weight $\beta$ and $s(\cdot)$, which is a function of staleness for adaptive $\beta_t$. We also list the frequently used symbols in this article in Tab. \ref{notations}.  Previous works and our experimental results show that Fedasync often fails to converge to the target accuracy due to the high data heterogeneity.

\begin{equation}
\left\{
\begin{array}{lr}
w_g^{t+1} = \left(1-\beta_t\right) w_g^t +\beta_t w_l^t \\
%& \\
\beta_t = \beta \times s(t - \tau) \\
\end{array}
\right.
\label{fedasync}
\end{equation}

\subsection{Security and Privacy Protection}
SecAgg is a privacy-enhancing technique based on cryptographic primitives \cite{bonawitz2016practical,so2021turbo} or hardware-based Trusted Execution Environment (TEE) \cite{cryptoeprint:2020/1561}. The server can only see the gradients after aggregation yet cannot know the original ones, which are private to each client. In this way, SecAgg ensures that the uploaded model is not accessible to the server and other malicious nodes or semi-honest nodes who follow the rules of engagement but are curious about the information of others. 

Since most cryptographic algorithms nowadays are based on homomorphic encryption, they cannot handle the situation in semi-asynchronous, where the parameters are aggregated from different versions each time, and asynchronous algorithms, where only one parameter is aggregated at a time. Nowadays there are also new encryption algorithms, such as BASecAgg\cite{so2021turbo}, that can be applied to semi-asynchronous algorithms to fulfill the requirements of secagg.

Regardless of whether cryptographic algorithms or TEEs are used, the basic requirement for these methods to be usable is that updates from one client can be hidden from updates from multiple clients.

\section{ \textit{FedFa}: Fully Asynchronous Federated Average for Federated Learning}
The fully asynchronous parameter update strategy has been promised to eliminate the waiting time and reduce the wall clock time for federated learning, while it also faces challenges for poor convergence performance and model accuracy degradation. In this section, we provide a detailed explanation for our \textit{FedFa}, a fully asynchronous parameter update strategy for federated learning that can eliminate the waiting time and guarantee the convergence of the federated learning.

\subsection{The Design of \textit{FedFa}}
The fully asynchronous parameter update strategy in the federated learning environment often leads to the performance oscillation of the global model on the server due to the heterogeneity of the data across different clients. Thus, to avoid this performance degradation, we propose merging the historical version of the model collected by the server into the most updated one that arrives in the server for parameter update. We define a \textit{Sliding Window} to buffer $K$ historical version of the model that arrived in the server and merge them into the current one. The workflow of the server and client for our proposed \textit{FedFa} is introduced in Algorithm \ref{algorithm_server} and \ref{algorithm_client}.

\begin{algorithm}[ht]
\footnotesize
\caption{\textit{FedFa}-Server}\label{algorithm_server}
\KwIn{Sliding window size $K$; Client learning rate $\eta_{l}$; Client steps $Q$, Buffer Queue $S$
}
\KwOut{Converged model}
initialize t = 0 , $w_g^{0}$;

Broadcast the initialized server model parameters to the clients which are selected in the first round; 

\Repeat{converge}{\If{get update $w_l^{t}$}{

\If{$t\leq K$}{S.Enqueue($w_l^{t}$) ;}\ 
\If{$t>K$}{S.Enqueue($w_l^{t}$) ;\\ S.Dequeue();\\ $w_g^{t+1} = \frac{1}{K} \sum_{k\in \mathcal{S}^t} w_k^{t -\tau_k(t)}$;}
\ t = t+1;}} 
\end{algorithm}

\begin{algorithm}[ht]
\footnotesize
\caption{\textit{FedFa}-Client}\label{algorithm_client}
 \KwIn{Client learning rate $\eta_{l}$; Client steps $Q$; Server model parameters $w_g$; Local parameters at $q_{th}$ steps $y_q$; Function to calculate gradient $g(\cdot)$
}
\KwOut{Client updated parameters $w_l$}   
$y_0 = w_g$; \\
$q = 0$;  \\
\While{$q<Q$}{$y_q = y_{q - 1} - \eta_lg(y_{q-1})$;\\ $q = q+1;$}
$w_l = y_Q ;$\\
send $w_l$ to server;
\end{algorithm}

% \begin{figure}[ht]
%   \vspace{-3pt}
% \captionsetup{font=small}
% 	\centering
% 	\includesvg[width=0.8\linewidth, height=0.45\linewidth]{images/sliding window.svg}
%   \vspace{-6pt}
% 	\caption{The wall clock time comparison.}
% 	\label{comm_time}
%  \vspace{-8pt}
% \end{figure}

During training, once the number of updates received in the buffer equals $K$ on the server, the regular training phase begins. When the buffer receives an update, the oldest update is removed so that the buffer always maintains the size of $K$. The parameters of each round on the server side are equal to the arithmetic mean of all parameters in the buffer. Aggregation is performed in a sliding window-like manner. If we do not perform this sliding window aggregation but wait for all the updates in the buffer to be replaced with newly received updates for each communication round, this is exactly equivalent to the FedBuff with buffer size $K$. Instead, we use sliding windows for aggregation, tapping into some intermediate states between the two rounds to approach the global optimization goal quickly.

\subsection{Gradient OR Parameter Transmission}
In the federated learning environment, exchanging the parameters and gradients between the server and clients are two common approaches for parameter updates. In synchronous parameter update strategies, transmission parameters are equivalent to transmission gradients or accumulated gradients. This can be obtained from the following formula (\ref{para_grad}). In synchronous parameter update strategies, the server receives the updated parameters as $w_k^t = w_g^{t-1} + \triangle_k^t$, where the $\triangle_k^t$ is the accumulated gradient in client $k$, and uses it to update the model based on the following formula (\ref{para_grad}).

\begin{equation}
\begin{aligned}
w_g^t &= \tfrac{1}{K} \sum_{k \in S_t} w_k^t = \tfrac{1}{K} \sum_{k \in S_t} (w_g^{t-1} + \triangle_k^t) \\
%& = \tfrac{1}{K} \sum_{k \in S_t} (w_g^{t-1} + \triangle_k^t)\\
&=  w_g^{t-1} + \tfrac{1}{K} \sum_{k \in S_t} \triangle_k^t 
\end{aligned}  
\label{para_grad}
\end{equation}

However, in asynchronous and semi-asynchronous parameter update strategies, the parameter update is calculated by $w_k^t = w_g^{t-\tau_k^t} + \triangle_k^t$. If parameters are transmitted between the server and clients, the parameter update of $w_k^t = w_g^{t-\tau_k^t} + \triangle_k^t$ is not equivalent to the above formula (\ref{para_grad}) due to the different phases at the start of training for each weight participating in the aggregation.

We implement and evaluate our \textit{FedFa} based on both the parameters transmission and the difference transmission strategy, called \textit{FedFa-Param} and \textit{FedFa-Delta}, respectively.

\section{Convergence Analysis of \textit{FedFa}}

\subsection{Problem Formulation}
The iterative formulation of \textit{FedFa-Param} and \textit{FedFa-Delta} can be outlined by equation (\ref{convergence1}) and equation (\ref{convergence2}), respectively. Both of them can achieve the same convergence rate as FedBuff \cite{nguyen2022federated} as long as the following 5 assumptions hold.

\begin{equation}
w_g^{t+1} = \frac{1}{K} \sum_{k\in \mathcal{S}^t}\left( w_k^{t -\tau_k(t)}\right)
\label{convergence1}
\end{equation}

\begin{equation}
w_g^{t+1}= w_g^{t} + \frac{1}{K} \sum_{k \in \mathcal{S}^t}\left(\triangle_k^{t-\tau_k(t)}\right)
\label{convergence2}
\end{equation}

\setlength{\parindent}{0pt} \textbf{Assumptions:} (without loss of generality)

\begin{itemize}
\item 1. \textit{Unbiasedness of client stochastic gradient}
     \begin{equation}
     \mathbb{E}_{\zeta_k}[g_k(w ; \zeta_k))]=\nabla F_k(w)
     \end{equation}
\item 2. \textit{Bounded local and global variance for all clients}
    \begin{equation}
     \mathbb{E}_{\zeta_k \mid k}\left[\left\|g_k\left(w ; \zeta_k\right)-\nabla F_k(w)\right\|^2\right] \leq \sigma_{\ell}^2
    \end{equation}
\item 3. \textit{Bounded gradient} $\left\|\nabla F_k\right\|^2 \leq G,  k \in [m]$

\item 4. \textit{Lipschitz gradient for all client $ k \in [m]$ and the gradient is L-smooth}
\begin{equation}
\left\|\nabla F_k(w)-\nabla F_k\left(w^{\prime}\right)\right\|^2 \leq L\left\|w-w^{\prime}\right\|^2
\end{equation}

\item 5. \textit{Bounded Staleness.} We use $\tau_k(t)$ to denote the staleness of client $k$ where we assume the server is in round $t$, and $\tau_k(t)$ has an upper bound $\tau_{max}$: 
\begin{equation}
1\leq\tau_k(t)\leq\tau_{max}
\end{equation}
\end{itemize}

\subsection{The Proof for Convergence Rate of \textit{FedFa}}
Inspired by FedBuff \cite{nguyen2022federated}, we provide an analysis of the convergence rate for \textit{FedFa} in this section. Several previous works also analyze the convergence of synchronous aggregation strategies \cite{li2020federated,stich2018local,wang2020tackling,zhao2018federated}, and semi-asynchronous strategies \cite{nguyen2022federated,mania2015perturbed,fraboni2023general,koloskova2022sharper}. Based on the above assumptions, the convergence rate of \textit{FedFa} can be calculated as formula (\ref{convergence_speed}),
where the $\frac{1}{T} \sum_{t=0}^{T-1}\left\|\nabla f\left(w^{t}\right)\right\|^2$ indicates the convergence rate. The detailed proof of the convergence rate can be found in the Appendix.

\textbf{Theorem 1}
\begin{equation}
\begin{aligned}
    % &\frac{1}{T} \sum_{t=\tau_{max}}^{T+\tau_{max}-1}\left\|\nabla f\left(w^{t-\tau_{max}}\right)\right\|^2 \leq\frac{2\left(f\left(w^0\right)-f\left(w^*\right)\right)}{\eta_g \alpha(Q) T}\\
    % & +3 L^2 Q \beta(Q)\left(\eta_g^2 \tau_{\max , K}^2+1\right)\left(\sigma_{\ell}^2
    % +\sigma_g^2+G\right)\\
    % &+\frac{L}{2} \frac{\eta_g \beta(Q)}{\alpha(Q)} \sigma_{\ell}^2
     &\frac{1}{T} \sum_{t=0}^{T-1}\left\|\nabla f\left(w_g^{t}\right)\right\|^2 \leq\frac{2\left(f\left(w^0\right)-f\left(w^*\right)\right)}{\eta_g \sum_{q=0}^{Q-1} \eta_{\ell}^{(q)} T}\\
    & +3 L^2 Q \sum_{q=0}^{Q-1}\left(\eta_{\ell}^{(q)}\right)^2\left(\eta_g^2 \tau_{\max}^2+1\right)\left(\sigma_{\ell}^2
    +\sigma_g^2+G\right)\\
    &+\frac{L}{2} \frac{\eta_g \sum_{q=0}^{Q-1}\left(\eta_{\ell}^{(q)}\right)^2}{\sum_{q=0}^{Q-1} \eta_{\ell}^{(q)}} \sigma_{\ell}^2
\end{aligned}
\label{convergence_speed}
\end{equation}

%Some of these parameters are given a value that matches our needs.
\textbf{Corollary.} Final convergence rate is obtained as formula (\ref{convergence22}).

\begin{equation}
    \begin{aligned}
& \frac{1}{T} \sum_{t=0}^{T-1} \mathbb{E}\left[\left\|\nabla f\left(w_g^t\right)\right\|^2\right] \leq \mathcal{O}\left(\frac{F^*}{\sqrt{T Q}}\right) \\
& +\mathcal{O}\left(\frac{\sigma_{\ell}^2}{\sqrt{T Q}}\right)+\mathcal{O}\left(\frac{Q \sigma^2}{T K^2}\right)+\mathcal{O}\left(\frac{Q \sigma^2 \tau_{\max}^2}{T K^2}\right)
\end{aligned}
\label{convergence22}
\end{equation}
where $F^*:=f\left(w^0\right)-f^*$, $\sigma^2:=\sigma_{\ell}^2+\sigma_g^2+G$, $\eta_{\ell}=\mathcal{O}(1 /(K \sqrt{T Q}))$, $\eta_g=\mathcal{O}(K)$.

\begin{figure*}[!t]
	\centering
	\subfloat[$\alpha=0.1$]{\includegraphics[width=0.3\linewidth, height=0.2\linewidth]{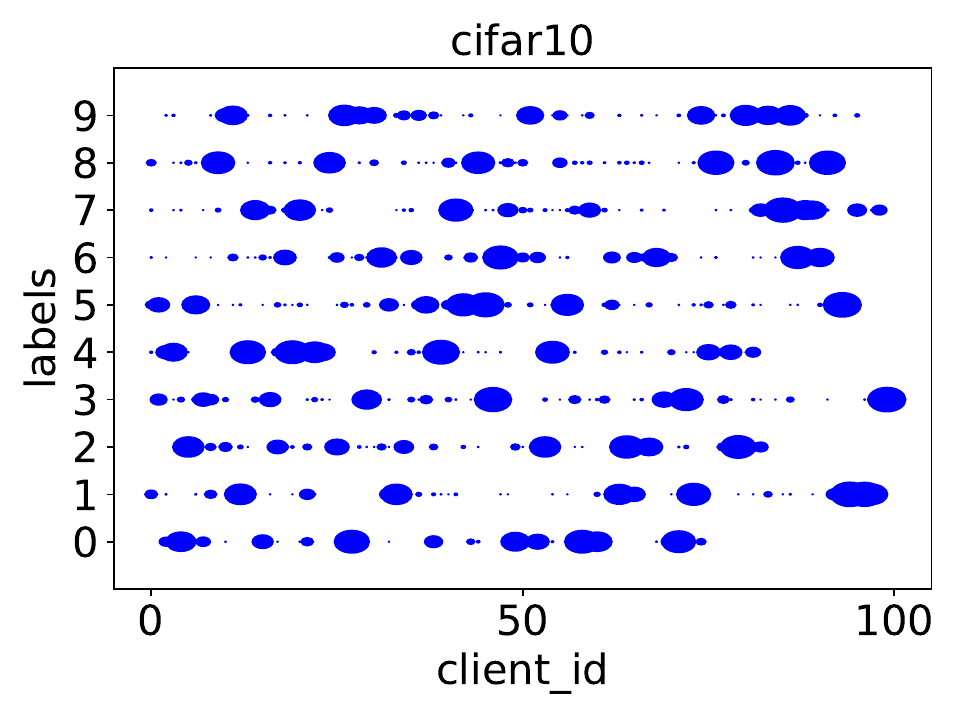}}
    \hfill
	\subfloat[$\alpha=0.5$]{\includegraphics[width=0.3\linewidth, height=0.2\linewidth]{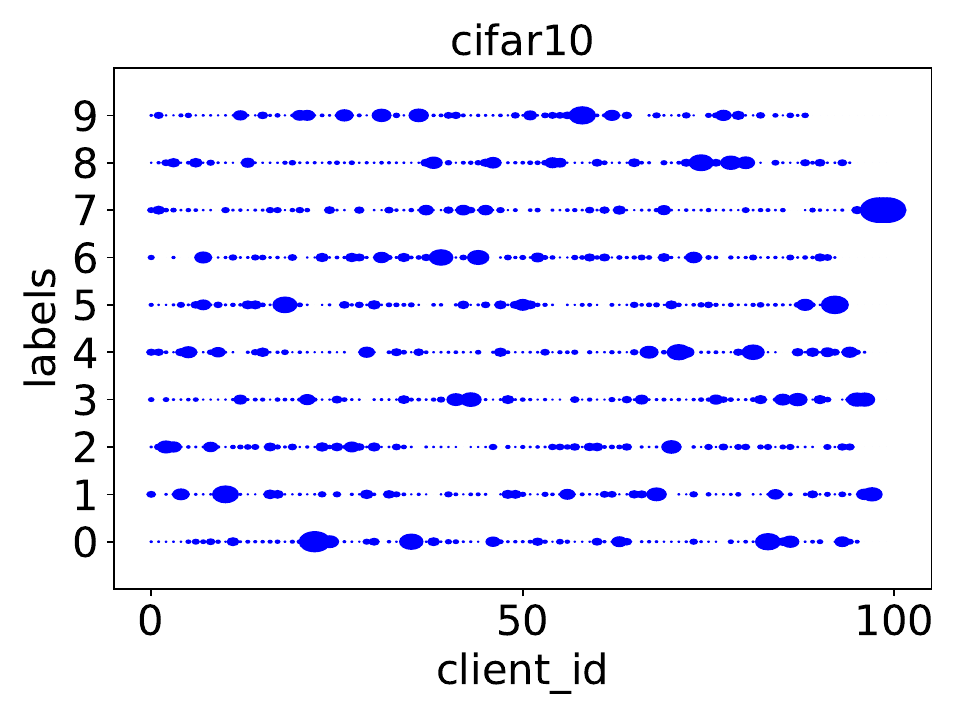}} 
   \hfill
   \subfloat[$\alpha=5$]{\includegraphics[width=0.3\linewidth, height=0.2\linewidth]{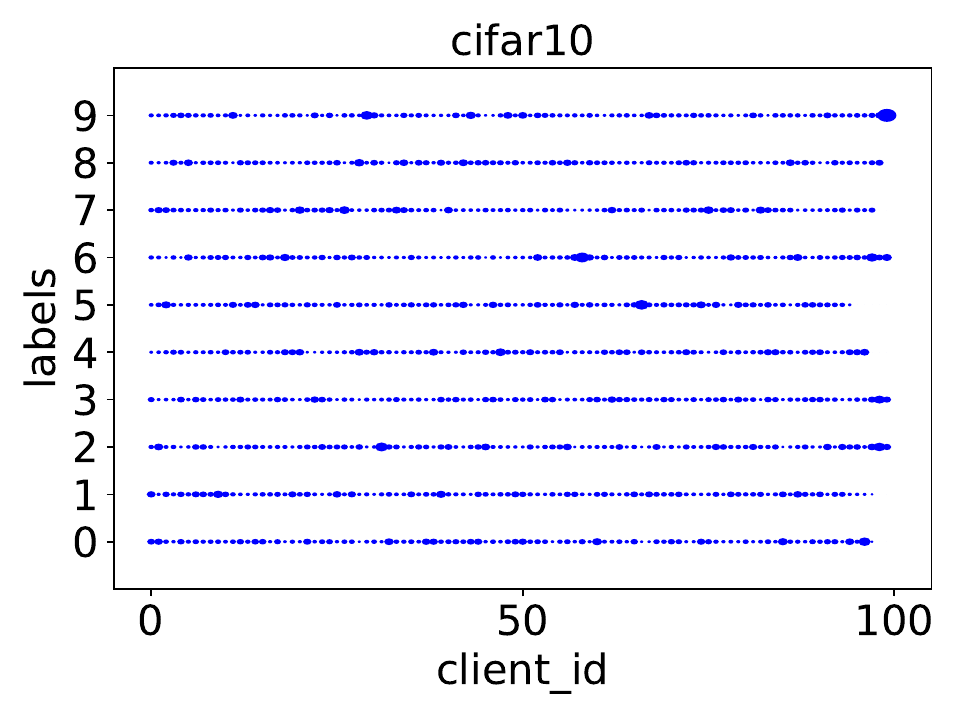}}\\ 
   \vspace{-6pt}
	\caption{Data distribution where a larger circle indicates a larger dataset size for each label 0-9 of CIFAR10.}
 \vspace{-15pt}
	\label{dirichlet}
\end{figure*}

The worst-case iteration complexity is captured by Corollary, which provides an upper bound on the ergodic norm-squared of the gradient—an essential metric in the study of non-convex stochastic optimization. If this norm-squared diminishes as the number of iterations (T) increases, it indicates that the norm-squared of the gradient at later iterations is approaching zero, suggesting convergence towards a first-order stationary point. Corollary 1 stems from Theorem 1 by employing a particular constant learning rate. It provides insights into the trade-offs involving loss convergence, local and global variance, the impact of client drift due to local steps, the influence of staleness, and the effect of buffer size.

We can conclude that our both version of \textit{FedFa} has the same convergence rate as Fedbuff statistically and theoretically.

\vspace{-6pt}
\section{Performance Evaluation and Analysis}
\subsection{Prototype Implementation}
We implement \textit{FedFa} on top of \textit{Plota} \cite{su2022asynchronous} and \textit{Pytorch}. Plota is a federated learning framework that supports temporal simulation for asynchronous federated learning on a single machine. Most of our experiments were performed on an NVIDIA GeForce RTX 3090 graphic card. For the task of fine-tuning the large language model, we performed it on 2× NVIDIA GeForce RTX 4090 graphic cards.

% \cite{Paszke_PyTorch_An_Imperative_2019}

\subsection{Evaluation Methodology}

\textbf{Benchmarks.} 
For the CV task, we compare our method \textit{FedFa} with other baselines on the most popular CV dataset, cifar10\cite{krizhevsky2009learning}. For the NLP task, we perform full-parameter fine-tuning on the sent140 \cite{caldas2018leaf} task and on the STT2 task using the parameter-efficient fine-tuning method LoRA \cite{hu2021lora}. For these three tasks mentioned above, we perform experiments using the ResNet18 \cite{he2016deep}, pre-trained Language Model Tiny-Bert \cite{bhargava2021generalization}, and the pre-trained Language Model bert-base model, respectively. To simulate the Non-IID environments, we partition the whole training data based on the Dirichlet Distribution and use a coefficient $\alpha$ to control the heterogeneity of the data distribution across clients \cite{hsu2019measuring}, where a small $\alpha$ represents higher data heterogeneity among each client, shown in Fig. \ref{dirichlet}. We subject the datasets Cifar10 and STT2 to the above division method that satisfies the Non-IID setting. The Sent140 is a dataset of sentiment categories collected from Twitter, which is naturally divided into federal settings by treating each user as a client, where we choose users with at least $100$ samples.

\setlength{\parindent}{0pt} \textbf{Experiment setup.} We deploy 100 clients and sample 10 clients for each communication round for ResNet18 trained on Cifar10. In the case of Tiny-Bert trained on Sent140, the number of clients is determined by the actual number of users in the dataset (i.e. 146) while maintaining 10 clients for training simultaneously. There was no manual setting for the number of clients. For the Bert-base model fine-tuned on the STT2 dataset, we deploy 10 clients and maintain 5 clients performing the training task simultaneously.

\textbf{Simulation of delay time.} In the experiments, each client needs to wait for a delay time $t$ before communicating with the server for data transmission after completing the training task. This delay time $t$ is simulated by the server and sent to each client. We set the simulated delay time for all clients with a long-tailed distribution. There are more clients with faster training and communication speeds and fewer clients with slower speeds.

\setlength{\parindent}{0pt} \textbf{Hyperparameters.} For the ResNet18 on Cifar10, the learning rate $\eta_g$ is set as $0.01$ with local epoch $E= 10$  and local mini-batches $lb = 32$. For the Tiny-Bert experiments on the Sent140 dataset, we set $\eta_g = 0.0004, E = 15, lb = 5$, which is inspired by \cite{cho2022heterogeneous}. For fine-tuning Bert on the STT2 dataset, we set $\eta_g = 1e-4, E = 1, lb = 32$. For the hyperparameter in LoRA settings, we set r = 1,$\alpha_{\text {LoRA }}=1$. 

\textbf{Baselines.} We compare our \textit{FedFa} with five other related approaches: (i) FedAvg \cite{mcmahan2017communication}, which is an approach in synchronous federated learning; (ii) FedBuff \cite{nguyen2022federated}, which is a semi-asynchronous approach for federated learning; (iii) FedAsync, the vanilla asynchronous approach for federated learning; (iv) Port \cite{su2022asynchronous}, which is also a semi-asynchronous approach.

% \begin{figure*}[!t]
% \captionsetup{font=small}
% %\vspace{-12pt}
%  \subfloat[Language Model on Sent140, Non-IID]{
%  \includegraphics[scale=0.28]{images/sent140-tinybert.pdf}
%   %\includegraphics[scale=0.23]{images/sent140-tinybert-1.pdf}
%  \includegraphics[scale=0.28]{images/sent140-tinybert-2.pdf}
%  \includegraphics[scale=0.28]{images/sent140-tinybert-3.pdf}
%  \includegraphics[scale=0.28]{images/sent140-tinybert-4.pdf}}
%  \vspace{3pt}
% % scale=0.35
% \subfloat[ResNet18 on Cifar10 where $\alpha = 0.1$, Non-IID]{
% \includegraphics[scale=0.28]{images/fed-cifar10.pdf}
% % \includegraphics[scale=0.23]{images/fed-cifar10-1.pdf}
% \includegraphics[scale=0.28]{images/fed-cifar10-2.pdf}
% \includegraphics[scale=0.28]{images/fed-cifar10-3.pdf}
% \includegraphics[scale=0.28]{images/fed-cifar10-4.pdf}
% }
% \vspace{3pt}b
% \subfloat[ResNet18 on Cifar10 where $\alpha = 5$, IID]{
% \includegraphics[scale=0.28]{images/fed-cifar10-a5.pdf}
% % \includegraphics[scale=0.23]{images/fed-cifar10-a5-1.pdf}
% \includegraphics[scale=0.28]{images/fed-cifar10-a5-2.pdf}
% \includegraphics[scale=0.28]{images/fed-cifar10-a5-3.pdf}
% \includegraphics[scale=0.28]{images/fed-cifar10-a5-4.pdf}
% }
% \vspace{-6pt}
% \caption{The performance comparison of different federated learning synchronization algorithms.}
% \label{comparision}
% \vspace{-8pt}
% \end{figure*}

\subsection{Results and Analysis}

\begin{figure}[!t]
% \vspace{-10pt}
	\centering
	\subfloat[$\alpha=0.1$]{\includegraphics[scale=0.28]{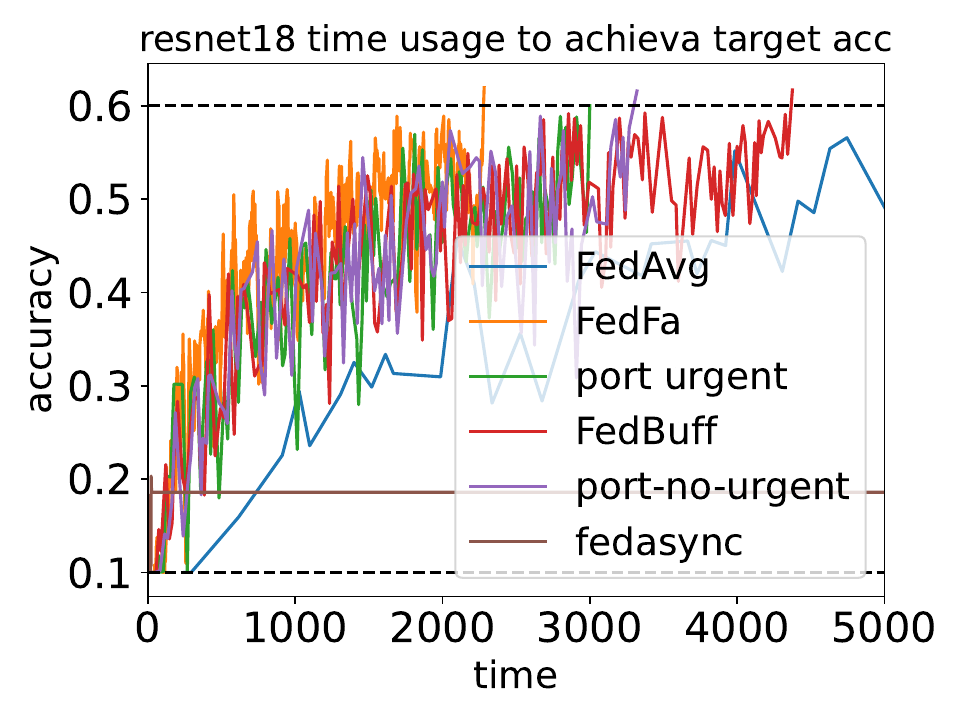}}
    % \hfill
	\subfloat[$\alpha=5$]{\includegraphics[scale=0.28]{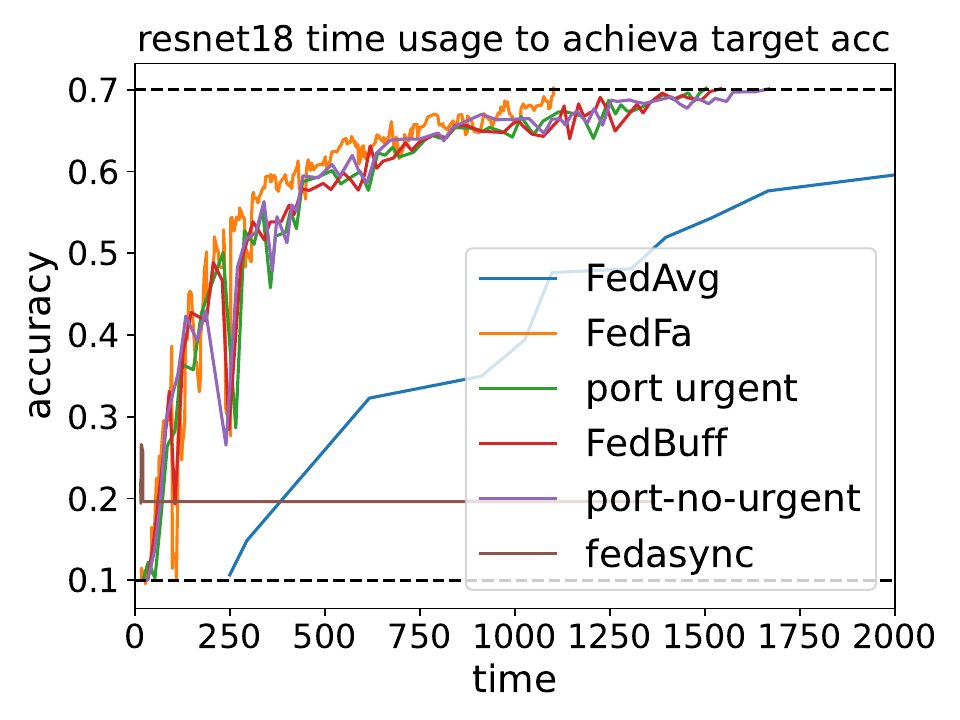}} \\ 
    \vspace{-8pt}
	\caption{Performance comparison of different training strategies on both IID and Non-IID data settings regarding the wall clock time.}
 \vspace{-12pt}
	\label{time-efff}
\end{figure}

\textbf{Time Efficiency of \textit{FedFa}.}
In this section, we analyze the validation accuracy in terms of the wall clock time based on different benchmarks that are shown in Fig. \ref{time-efff}(a)(b). These figures show that \textit{FedFa} can achieve the same accuracy as semi-asynchronous and synchronous parameter update strategies using less time. Furthermore, the previous Fedasync strategy cannot converge to the target accuracy. We also give detailed results about the wall clock time to achieve the target accuracy for different benchmarks in Tab. \ref{acc-time}. We find that in terms of time efficiency, the fully asynchronous algorithm is more efficient than the semi-asynchronous and synchronous algorithms due to the natural effects of the system design. The Fedasync algorithm fails to achieve the target accuracy. 

To further evaluate the efficiency of our proposed \textit{FedFa}, we also conduct experiments based on the parameter update strategy, called \textit{FedFa-Delta}, in which clients send the parameter difference to the server, and the server sends the updated parameters to clients for each communication round. The parameter update strategy in which clients send parameters to the server and the server sends updated parameters to clients is called \textit{FedFa-Para}. From Tab. \ref{acc-time}, we observe that both of the two versions of our method outperform existing approaches, especially for \textit{FedFa-Delta}. \textit{FedFa-Delta} achieves up to $7\times$ and $5\times$ speedup compared to synchronous methods and semi-asynchronous methods, respectively. At the same time, from Fig. \ref{time-efff} and Tab. \ref{acc-time}, we can see that Fedasync does not achieve the target accuracy, which also shows that our method can not only improve the time efficiency but also stabilize the training process.

%Of the two versions of \textit{FedFa}, the version that transmits the difference, also known as \textit{FedFa-delta}, is relatively more efficient. For the semi-asynchronous method, our method gets an acceleration ratio ranging from about 2× to 5×. This also shows that the actual time efficiency of the semi-asynchronous method is not necessarily much higher than the synchronous method. 

\begin{table}[ht]
\footnotesize		
		\label{tab:da16}
  \vspace{-5pt}
  \caption{Wall Clock Time for Target Accuracy}
  \vspace{-5pt}
	\centering 
         \setlength{\tabcolsep}{0.5mm}{
		\begin{tabular}{ccccc} 
			\toprule[1.5pt]
		 Dataset & \multicolumn{1}{c}{Sent140}& \multicolumn{1}{c}{STT2}& \multicolumn{2}{c}{Cifar10}  \\
		\cmidrule(lr){4-5}
  		 & Non-IID & Non-IID&Non-IID & IID   \\
        \midrule[1pt]
    		Target-Acc  & 0.71 &0.85&  0.6 &  0.7   \\
            \midrule[1pt]
		FedAvg  & 1139(6$\times$) &4660(7$\times$)& 9833(5$\times$) & 5040(6$\times$)   \\
		Fedasync  & failed & failed &failed& failed   \\
		FedBuff  & 485(3$\times$) &2985(4.5$\times$)&  4375(2$\times$) & 1541(2$\times$)   \\
		Port-Urgent  & 767(5$\times$) &2325(3.5$\times$)& 3000(1.6$\times$) & 1503(1.7$\times$)  \\
		Port-Non-Urgent  & 662(4$\times$) &2770(4.2$\times$)& 3320(1.7$\times$) & 1668(2$\times$)  \\
		FedFa-Param  & 191(1.1$\times$) &829(1.2$\times$)& 2282(1.2$\times$) & 1102(1.3$\times$)  \\
        FedFa-Delta  & \textbf{\underline{165}} &\textbf{\underline{652}}& \textbf{\underline{1917}} & \textbf{\underline{855}}  \\
			\bottomrule[1.5pt] 
		\end{tabular}}
      
   \label{acc-time}

\end{table}

\begin{figure}[!t]
% \vspace{-10pt}
	\centering
	\subfloat[$\alpha=0.1$]{\includegraphics[scale=0.28]{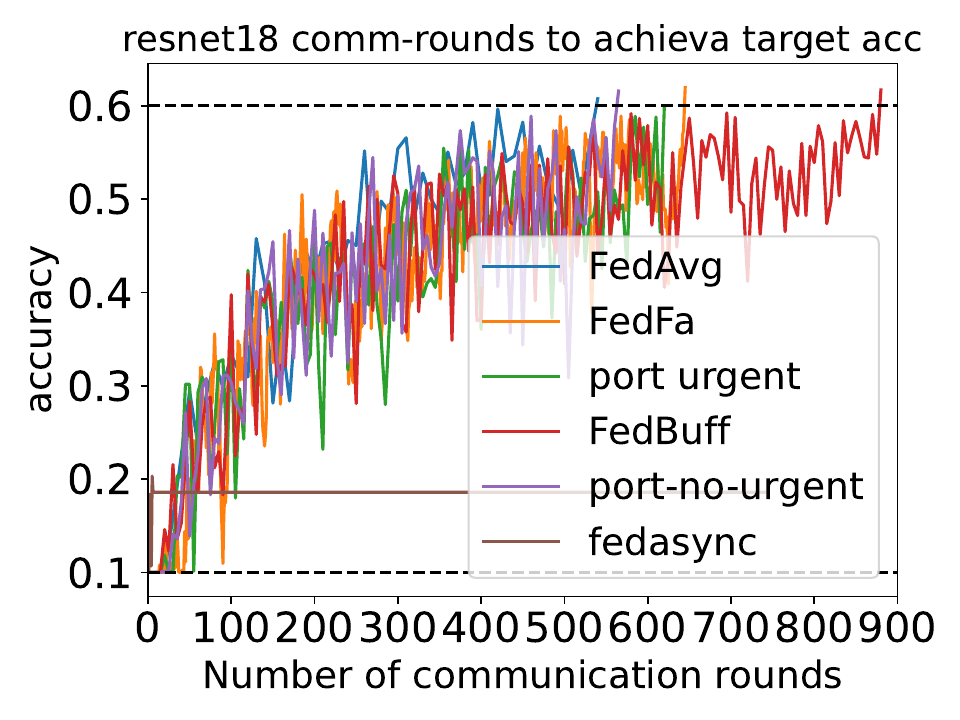}}
    % \hfill
	\subfloat[$\alpha=5$]{\includegraphics[scale=0.28]{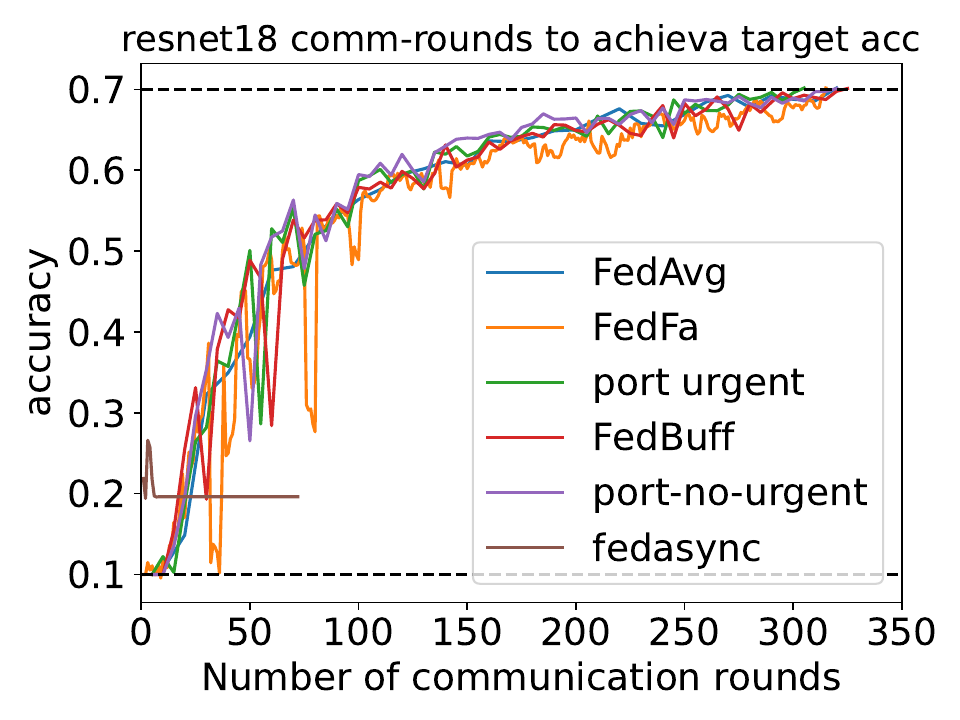}} \\ 
    \vspace{-8pt}
	\caption{Performance comparison of different training strategies on both IID and Non-IID data settings regarding communication round.}
  \vspace{-12pt}
	\label{comm-efff}
\end{figure}

\textbf{Communication Efficiency of \textit{FedFa}.}
In this section, we analyze the validation accuracy in terms of the communication rounds based on different benchmarks that are shown in Fig. \ref{comm-efff}. We also provide the communication rounds required to achieve the target accuracy for different benchmarks in Tab. \ref{acc-comm}. Typically, more communication rounds mean a longer period of time is used in the federated learning training process. However, since semi-asynchronous and fully asynchronous federated learning algorithms reduce and eliminate the waiting time to some extent, these algorithms, while time efficient, may require more communication rounds to reach the target accuracy. In Tab. \ref{acc-comm}, we observe that our proposed \textit{FedFa-Delta} outperforms other existing approaches by up to $3\times$. Combining the wall clock time that is shown in Tab. \ref{acc-time}, the \textit{FedFa-Delta} also outperforms the existing approaches for the whole training time.

%The \textit{FedFa-Param} in Tab. \ref{acc-time} is second only to our \textit{FedFa-delta} algorithm in terms of time efficiency. Still, it is not one of the top two in terms of the number of communication rounds. In this respect, the difference between synchronous, semi-asynchronous, and asynchronous algorithms is insignificant. Our \textit{FedFa-delta} algorithm still causes the most minor communication consumption and is the most communication efficient among all benchmarks. It can provide a speedup ratio of 1.2-3 compared to other algorithms.

% This table shows that \textit{FedFa} can achieve the target accuracy using similar communication rounds as other semi-asynchronous and synchronous strategies, while \textit{FedFa} can reduce the wall clock time for each communication round significantly introduced in the section \textit{Time Efficiency of \textit{FedFa}}. Thus, the overall time consumption to achieve the target accuracy of \textit{FedFa} is much less than others.

\begin{table}[ht]
\footnotesize		
		\label{tab:da16}
  \vspace{-5pt}
  \caption{ Communication Rounds for Target Accuracy}
  \vspace{-5pt}
	\centering  
         \setlength{\tabcolsep}{1.5mm}{
		\begin{tabular}{ccccc} 
			\toprule[1.5pt]
		 Dataset & \multicolumn{1}{c}{Sent140}& \multicolumn{1}{c}{STT2}& \multicolumn{2}{c}{Cifar10}  \\
		\cmidrule(lr){4-5}
  		 & Non-IID &Non-IID &Non-IID & IID   \\
     \midrule[1pt]
    		Target-Acc  & 0.71 &0.85&  0.6 &  0.7   \\
            \midrule[1pt]
    		FedAvg  & 70(2$\times$) &120(1.6$\times$)& 540(1.2$\times$) & 320 (1.4$\times$) \\
    		Fedasync  & failed & failed&failed&  failed  \\
    		FedBuff  & 70(2$\times$) &123(1.6$\times$)& 880(1.9$\times$) & 325(1.4$\times$)  \\
    		Port-Urgent  & 110(3$\times$) &98(1.3$\times$)& 620(1.3$\times$) & 305(1.3$\times$) \\
    		Port-Non-Urgent  & 95(2$\times$) &112(1.5$\times$)& 565(1.2$\times$) & 320(1.4$\times$) \\
    		FedFa-Param  & 45(1.3$\times$) &133(2$\times$)& 645(1.4$\times$) & 315(1.4$\times$) \\
                FedFa-Delta & \textbf{\underline{36}}& \textbf{\underline{73}}& \textbf{\underline{465}} & \textbf{\underline{230}} \\
			\bottomrule[1.5pt] 
		\end{tabular}} 
  \vspace{-10pt}
   \label{acc-comm}
\end{table}

\textbf{Best Accuracy.}
Tab. \ref{best-acc} shows the optimal accuracy of the three foundational paradigms within a fixed number of communication rounds. We observe that \textit{FedFa-Delta} achieves the highest accuracy for both Sent140 and Cifar10-IID, and FedAvg achieves the highest accuracy for Cifar10-NonIID. The accuracy of all three paradigms is similar to each other, while our fully asynchronous approach \textit{FedFa-Delta} is time efficient. Thus, \textit{FedFa} is more suitable for scaling up the training scales in federated learning by eliminating the waiting time (without the barrier setting) compared to the synchronous and semi-asynchronous strategies.

\begin{table}[ht]
    \vspace{-6pt}
   \centering
\footnotesize
     \caption{ Best Accuracy Comparison} 		
    \vspace{-6pt}
		\label{tab:da16}
		\centering  
         \setlength{\tabcolsep}{1mm}{
		\begin{tabular}{ccccc} 
			\toprule[1.5pt]
			 		  Dataset & \multicolumn{1}{c}{Sent140}& \multicolumn{1}{c}{STT2}& \multicolumn{2}{c}{Cifar10}  \\
		\cmidrule(lr){4-5}
  		 & Non-IID &Non-IID &Non-IID & IID   \\
            \midrule[1pt]
    		FedAvg  & 0.7209 &   \textbf{\underline{0.891}}& \textbf{\underline{0.6570}} & 0.7690 \\
    		% Fedasync  &  &  &   \\
    		FedBuff  & 0.7209 & 0.871   &0.6557  &   0.7776\\
    		FedFa-Param  & 0.7291 &  0.88 & 0.6539 & 0.7781 \\
                FedFa-Delta & \textbf{\underline{0.7403}} &  0.883 & 0.6474 & \textbf{\underline{0.7831}} \\
			\bottomrule[1.5pt] 
		\end{tabular}}
   \label{best-acc}
      % \vspace{-6pt}
\end{table}

\begin{figure}[ht]
\captionsetup{}
	\centering
 \includegraphics[scale=0.42]{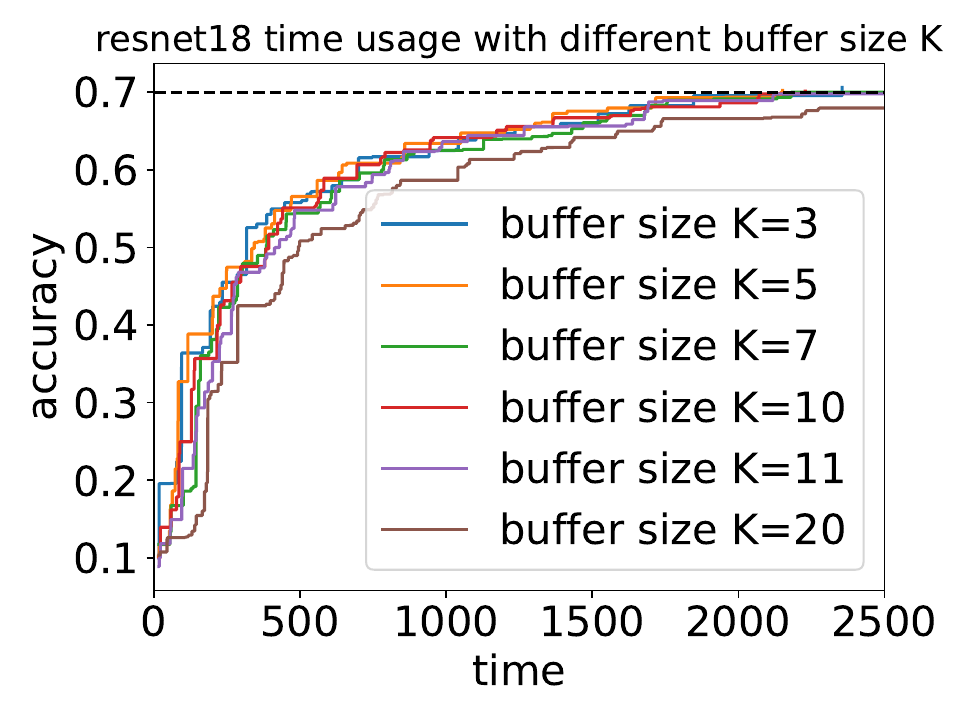} 
	\caption{The comparison of different buffer sizes $K$.}
	\label{bufferk}
\end{figure}

\textbf{Effective of the Buffer Size $K$.}
Fig. \ref{bufferk} shows the current optimal model accuracy ladder diagram of \textit{FedFa-Param} with different buffer sizes $K$. Changing the buffer size $K$ within a reasonable range does not significantly impact the model accuracy, while vastly increasing the $K$ will slow down the training procedure, such as setting the $K=20$. This is reasonable because it will excessively average the outdated information during each aggregation when setting the $K=20$. These results indicate that our proposed \textit{FedFa} is robust to the buffer size $K$.

% In our proposed \textit{FedFa}, we sample $M_c$ clients and collect results from only one client for each training round. This is different to \textit{FedAvg}.
\textbf{Effective of concurrency $M_c$}The number of clients sampled at the beginning of the federated learning training is called the number of concurrency $M_c$. Whenever K clients complete the training task and upload the parameters to the server, other K clients are resampled to maintain the number of concurrency $M_c$. Therefore, it also means the number of simultaneously active clients. In most federated learning experiment setups, $M_c$ is set as $1/10$ to $1/5$ of the total number of clients. In asynchronous algorithms, increasing the number of sampling clients may lead to worse training results due to the fact that staleness is directly related to the number of clients that are sampled.
As shown in Tab. \ref{mc}, the results of all the algorithms show some decrease in model accuracy as $M_c$ increases. Moreover, our algorithm is less insensitive to $M_c$ than other algorithms.

%The $M_c$ indicates the number of clients that are sampled for each training round. 

\begin{table}[ht]
    \caption{The best accuracy of the model which is trained within a defined period of time in different scales of $M_c$} 
   \centering
\footnotesize		
   % \vspace{-6pt}
		\label{tab:da16}
		\centering
  
         \setlength{\tabcolsep}{2.5mm}{
		\begin{tabular}{ccccc} 
			\toprule[1.5pt]
			 		 $M_c$ & 2 & 5 & 10&20 \\
%                      &         & non-iid & iid   \\
            \midrule[1pt]
    		FedAvg  & 0.7101 &0.6757 &0.6207  &0.6017\\
    		
    		FedBuff  &0.7137  &0.6688  & 0.6327  &0.6045\\
    		% FedFa-param  &  &  &  &\\
                FedFa-Delta & 0.718 & 0.6755 & 0.642 &0.6356\\
			\bottomrule[1.5pt] 
		\end{tabular}}
   \label{mc}
 
      % \vspace{-6pt}
\end{table}

% \begin{figure}[ht]
%  \vspace{-3pt}
% \captionsetup{font=small}
% 	\centering
%  \includegraphics[width=0.7\linewidth, height=0.45\linewidth]{./images/szie_k} 
%   \vspace{-6pt}
% 	\caption{The comparison of different buffer sizes $K$.}
% 	\label{bufferk}
%  \vspace{-8pt}
% \end{figure}

\textbf{Combined with Synchronous Optimization Methods.}

\textit{FedFa} is highly portable and extensible since it only involves a change in the aggregation paradigm. It is easy to extend other synchronous federated learning optimization methods into their versions under the fully asynchronous paradigm of \textit{FedFa}. Moreover, such an extension can also eliminate the waiting time and improve the efficiency of the federated learning training process.

\begin{figure}[ht]
\captionsetup{font=small}
	\centering
 \includegraphics[scale=0.42]{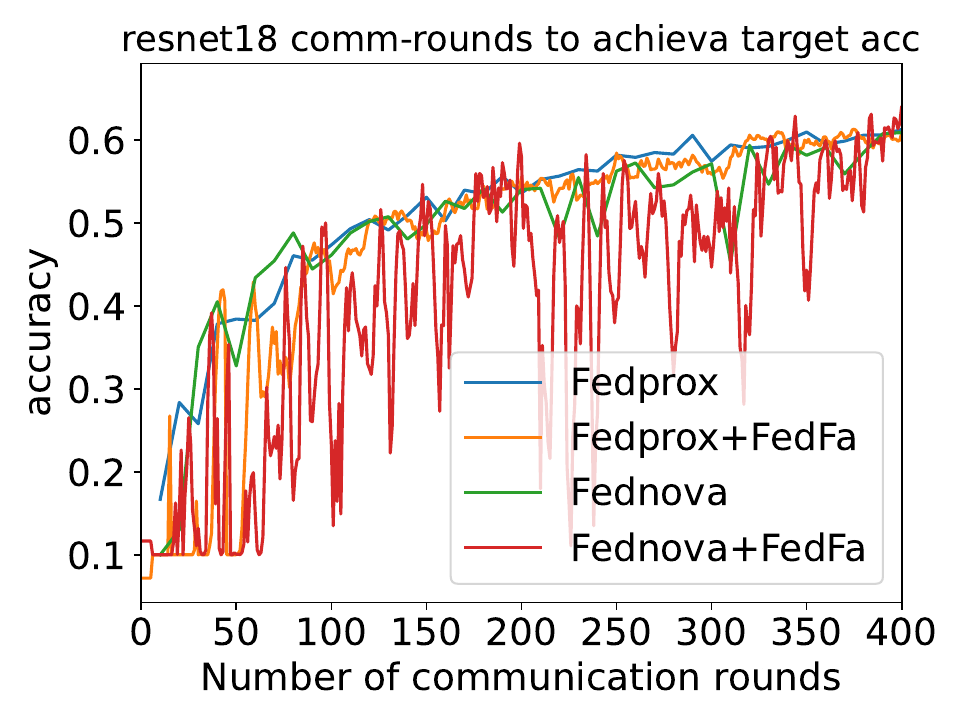} 
 \vspace{-6pt}
	\caption{Comparison of the model convergence between Original and FedFa versions of the optimization algorithm}
	\label{exnend-com}
 \vspace{-8pt}
\end{figure}

We integrate two synchronous federated learning optimization algorithms, \textit{Fedprox} \cite{li2020federated} and \textit{Fednova} \cite{wang2020tackling}, into our proposed \textit{FedFa-Delta}, which is a fully asynchronous paradigm. The synchronous federated learning versions of these two algorithms and the version under the full asynchronous paradigm, respectively, are used to perform the experiments on the Cifar10 dataset mentioned before. The number of communication rounds and the number of wall-clock times to achieve the target accuracy using these algorithms are shown in Tab. \ref{exten}. We observe that the wall-clock time for the whole federated learning training process is reduced to $1/3$ of the original time with only $1/10$ more communication rounds consumption. As shown in Fig. \ref{exnend-com}, the convergence of the \textit{FedFa} with \textit{Fedprox} is similar to the original algorithm. However, the training process of the \textit{FedFa} with \textit{Fednova} has become unstable. This is reasonable because the training process of asynchronous algorithms is inherently more unstable than synchronous algorithms.

\begin{table}[ht]
\caption{Comparison of communication rounds and wall clock time for target accuracy between optimization methods with (w/) and without (w/o) extending to the FedFa paradigm. } 
   \centering
\footnotesize
		\label{tab:da16}
		\centering
  
         \setlength{\tabcolsep}{2.5mm}{
            \begin{tabular}{ccc} 
			\toprule[1.5pt]
			 	Metrics	  & Fednova w/o &  Fedprox w/o \\
%                      &         & non-iid & iid   \\
            \midrule[1pt]

    		Comm-Rounds  & 331 / 390 &  320 / 290  \\
    		Wall Time  &  1290 / 4576 &  1184 / 3469 \\

			\bottomrule[1.5pt] 
		\end{tabular}}
   \label{exten}

      % \vspace{-6pt}
\end{table}

%\vspace{-6pt}
\section{Conclusion and Future Work}  
We propose \textit{FedFa}, a fully asynchronous parameter update strategy for federated learning. Unlike previous fully asynchronous algorithms, it satisfies the paradigm of secure aggregation and guarantees a stable and fast training process. Compared to the synchronous and semi-asynchronous algorithms, it improves the training performance considerably by up to $6\times$ and $4.0\times$ speed up for different benchmarks. Also, unlike other asynchronous algorithms, our approach provides some speedup in the number of communication rounds. We also give theoretical proof of the convergence rate of our proposed \textit{FedFa}, which has the same convergence upper bound as the widely used FedBuff. Finally, we also briefly implement the fully asynchronous version of some synchronous federated optimization algorithms in our \textit{FedFa} paradigm to make them more time efficient. We will explore more parameter update strategies for federated learning in the future. 

\section*{Acknowledgment}

This research was supported by the National Natural Science Foundation of China 62302420; U.S. Department of Energy, Office of Science, Advanced Scientific Computing Research (ASCR), under contract DE-AC02-06CH11357.

\clearpage
\bibliographystyle{named}
\bibliography{ijcai24}

\clearpage
\appendix
\textbf{Convergence Rate Proof}

\begin{equation}
\begin{aligned}
&w_g^{t+1} 
=\frac{1}{K} \sum_{k\in \mathcal{S}^t}\left( w_k^{t -\tau_k(t)}\right)\\
&=\frac{1}{K}\sum_{k\in \mathcal{S}^t}w_g^{t -\tau_k(t)}\\
&+\eta_g \frac{1}{K} \sum_{k \in \mathcal{S}^t}\left(-\eta_{\ell} \sum_{q=1}^Q g_k\left(y_{k, q}^{t-\tau_k(t)}\right)\right) \end{aligned}
\end{equation}
In the worst case scenario,$w_g^{t'}$ denotes the global weight that is least optimized in the buffer.In most cases, $t' = t - \tau_{max}$ 
\begin{equation}
    \begin{aligned}
    &w_g^{t+1} = \frac{1}{K} \sum_{k\in \mathcal{S}^t}\left( w_k^{t -\tau_k(t)}\right)\\
    &=\frac{1}{K}*Kw_g^{t'}+\eta_g \frac{1}{K} \sum_{k \in \mathcal{S}^t}\left(-\eta_{\ell} \sum_{q=1}^Q g_k\left(y_{k, q}^{t-\tau_k(t)}\right)\right)\\&=w_g^{t'}+\eta_g \frac{1}{K} \sum_{k \in \mathcal{S}^t}\left(-\eta_{\ell} \sum_{q=1}^Q g_k\left(y_{k, q}^{t-\tau_k(t)}\right)\right) 
\end{aligned}
\end{equation}

\textbf{Lemma 1.}
$\mathbb{E}\left[\left\|g_k\right\|^2\right] \leq 3\left(\sigma_{\ell}^2+\sigma_g^2+G\right)$
where the total expectation E[·] is evaluated over the randomness with
respect to client participation and the stochastic gradient taken by a client.
This has been proven in FedBuff.

By the L-smoothness assumption,
\begin{equation}
    \begin{aligned}
f\left(w^{t+1}\right) 
& \leq f\left(w^{t'}\right)-\eta_g\left\langle\nabla f\left(w^{t'}\right), \bar{\Delta}^t\right\rangle+\frac{L \eta_g^2}{2}\left\|\bar{\Delta}^t\right\|^2 \\
& \leq f\left(w^{t'}\right) \underbrace{-\frac{\eta_g}{K} \sum_{k \in \mathcal{S}_t}\left\langle\nabla f\left(w^{t'}\right), \Delta_k^{t-\tau_k}\right\rangle}_{T_1}\\
&+\underbrace{\frac{L \eta_g^2}{2 K^2}\left\|\sum_{k \in \mathcal{S}_t} \Delta_k^{t-\tau_k}\right\|^2}_{T_2}
\end{aligned}
\end{equation}

\begin{equation}
\begin{aligned}
     &T_1=-\frac{\eta_g}{K} \sum_{k \in \mathcal{S}_t}\left\langle\nabla f\left(w^{t'}\right), \sum_{q=0}^{Q-1} \eta_{\ell}^{(q)} g_k\left(y_{k, q}^{t-\tau_k}\right)\right\rangle\\
    &=-\frac{\eta_g}{K} \sum_{k \in \mathcal{S}_t} \sum_{q=0}^{Q-1} \eta_{\ell}^{(q)}\left\langle\nabla f\left(w^{t'}\right), g_k\left(y_{k, q}^{t-\tau_k}\right)\right\rangle
\end{aligned}  
\end{equation}

$\mathbb{E}[\cdot]:=\mathbb{E}_{\mathcal{H}} \mathbb{E}_{i \sim[m]} \mathbb{E}_{g_i \mid i, \mathcal{H}}[\cdot]$

\begin{equation}
    \begin{aligned}
        \begin{aligned}
&\mathbb{E}\left[T_1\right]  =-\mathbb{E}\left[\frac{\eta_g}{K} \sum_{k \in \mathcal{S}_t} \sum_{q=0}^{Q-1} \eta_{\ell}^{(q)}\left\langle\nabla f\left(w^{t'}\right), g_k\left(y_{k, q}^{t-\tau_k}\right)\right\rangle\right] \\
& =-\eta_g \mathbb{E}_{\mathcal{H}}\left[\frac{1}{m} \sum_{k=1}^m \sum_{q=0}^{Q-1} \eta_{\ell}^{(q)} \mathbb{E}_{g_k \mid k \sim[m]}\left\langle\nabla f\left(w^{t'}\right), g_k\left(y_{k, q}^{t-\tau_k}\right)\right\rangle\right] \\
& =-\frac{\eta_g}{m} \mathbb{E}_{\mathcal{H}}\left[\sum_{k=1}^m \sum_{q=0}^{Q-1} \eta_{\ell}^{(q)}\left\langle\nabla f\left(w^{t'}\right), \nabla F_k\left(y_{k, q}^{t-\tau_k}\right)\right\rangle\right] \\
& =-\eta_g \mathbb{E}_{\mathcal{H}}\left[\sum_{q=0}^{Q-1} \eta_{\ell}^{(q)}\left\langle\nabla f\left(w^{t'}\right), \frac{1}{m} \sum_{k=1}^m \nabla F_k\left(y_{k, q}^{t-\tau_k}\right)\right\rangle\right] .
\end{aligned}
    \end{aligned}
\end{equation}

$\langle a, b\rangle=\frac{1}{2}\left(\|a\|^2+\|b\|^2-\|a-b\|^2\right)$

\begin{equation}
    \begin{aligned}
\mathbb{E}\left[T_1\right] & =-\frac{\eta_g}{2}\left(\sum_{q=0}^{Q-1} \eta_{\ell}^{(q)}\right)\left\|\nabla f\left(w^{t'}\right)\right\|^2\\
&+\sum_{q=0}^{Q-1} \frac{\eta_g \eta_{\ell}^{(q)}}{2}\left(-\mathbb{E}_{\mathcal{H}}\left\|\frac{1}{m} \sum_{k=1}^m \nabla F_k\left(y_{k, q}^{t-\tau_k}\right)\right\|^2\right. \\
+ & \mathbb{E}_{\mathcal{H}} \underbrace{\left\|\nabla f\left(w^{t'}\right)-\frac{1}{m} \sum_{k=1}^m \nabla F_i\left(y_{k, q}^{t-\tau_k}\right)\right\|^2}_{T_3}) .
\end{aligned}
\end{equation}

 \begin{scriptsize}   

\begin{equation}
    \begin{aligned}
&\mathbb{E}\left[T_3\right]  = \\
&\frac{1}{m} \sum_{i=1}^m \mathbb{E}_{\mathcal{H}}\left\|\nabla F_k\left(w^{t'}\right)-\nabla F_k\left(w^{t-\tau_k}\right)
+ \nabla F_k\left(w^{t-\tau_k}\right)-\nabla F_k\left(y_{k, q}^{t-\tau_k}\right)\right\|^2 \\
&\leq \frac{2}{m} \sum_{k=1}^m \mathbb{E}_{\mathcal{H}}(\underbrace{\left\|\nabla F_k\left(w^{t'}\right)-\nabla F_k\left(w^{t-\tau_k}\right)\right\|^2}_{\text {staleness }}\\
&+ \underbrace{\left\|\nabla F_k\left(w^{t-\tau_k}\right)-\nabla F_k\left(y_{k, q}^{t-\tau_k}\right)\right\|^2}_{\text {local drift }}) \\
& \leq \frac{2}{m} \sum_{k=1}^m\left(L^2 \mathbb{E}_{\mathcal{H}}\left\|w^{t'}-w^{t-\tau_k}\right\|^2+L^2 \mathbb{E}_{\mathcal{H}}\left\|w^{t-\tau_k}-y_{k, q}^{t-\tau_k}\right\|^2\right) .
\end{aligned}
\end{equation}
\end{scriptsize}

\begin{equation}
    \begin{aligned}\left\|w^{t'}-w^{t-\tau_k}\right\|^2 
    & =\left\|\sum_{\rho=t'}^{t-\tau_k}\left(w^{\rho+1}-w_\rho\right)\right\|^2\\
    &=\left\|\sum_{\rho=t'}^{t-\tau_k} \frac{\eta_g}{K} \sum_{j_\rho \in \mathcal{S}_\rho} \Delta_{j_\rho}^\rho\right\|^2 \\
    & =\frac{\eta_g^2}{K^2}\left\|\sum_{\rho=t'}^{t-\tau_k} \sum_{j_\rho \in \mathcal{S}_\rho} \sum_{l=0}^{Q-1} \eta_{\ell}^{(l)} g_{j_\rho}\left(y_{j_\rho, l}^\rho\right)\right\|^2 .\end{aligned}
\end{equation}

Taking the expectation in terms of $\mathcal{H}$,
\begin{footnotesize}
\begin{equation}
    \begin{aligned}&\mathbb{E}_{\mathcal{H}}\left\|w^{t'}-w^{t-\tau_k}\right\|^2 \\ 
    &\leq 3 \eta_g^2 Q \max _{t - t' - \tau_i} {(t - t' - \tau_k)}^2\left(\sum_{l=0}^{Q-1}\left(\eta_{\ell}^{(l)}\right)^2\right)\left(\sigma_{\ell}^2+\sigma_g^2+G\right) \\
    &\leq 3 \eta_g^2 Q \max _{\tau_{max}-\tau_k} {(\tau_{max}-\tau_k)}^2\left(\sum_{l=0}^{Q-1}\left(\eta_{\ell}^{(l)}\right)^2\right)\left(\sigma_{\ell}^2+\sigma_g^2+G\right) \\ 
    &\leq 3 \eta_g^2 Q \tau_{\max}^2\left(\sum_{l=0}^{Q-1}\left(\eta_{\ell}^{(l)}\right)^2\right)\left(\sigma_{\ell}^2+\sigma_g^2+G\right)
    \end{aligned}
\end{equation}
\end{footnotesize}
\begin{equation}
    \begin{aligned}
        \mathbb{E}\left\|w^{t-\tau_k}-y_{k, q}^{t-\tau_k}\right\|^2
        &=\mathbb{E}\left\|y_{k, 0}^{t-\tau_i}-y_{k, q}^{t-\tau_k}\right\|^2 \\
        &\leq \mathbb{E}\left\|\sum_{l=0}^{q-1} \eta_{\ell}^{(l)} g_i\left(y_{k, l}^{t-\tau_k}\right)\right\|^2 \\
        &\leq 3 q\left(\sum_{l=0}^{q-1}\left(\eta_{\ell}^{(l)}\right)^2\right)\left(\sigma_{\ell}^2+\sigma_g^2+G\right)
    \end{aligned}
\end{equation}

\begin{footnotesize}

\begin{equation}
    \begin{aligned}
\mathbb{E}\left[T_3\right] 
& \leq 6L^2 \eta_g^2 Q \tau_{\max , K}^2\left(\sum_{k=0}^{Q-1}\left(\eta_{\ell}^{(k)}\right)^2\right)\left(\sigma_{\ell}^2+\sigma_g^2+G\right)\\
&+6L^2 q\left(\sum_{k=0}^{q-1}\left(\eta_{\ell}^{(k)}\right)^2\right)\left(\sigma_{\ell}^2+\sigma_g^2+G\right) \\
& \leq 6 L^2\left(\sum_{k=0}^{Q-1}\left(\eta_{\ell}^{(k)}\right)^2\right)\left(\eta_g^2 Q \tau_{\max}^2+q\right)\left(\sigma_{\ell}^2+\sigma_g^2+G\right) \\
& \leq 6 L^2 Q\left(\sum_{k=0}^{Q-1}\left(\eta_{\ell}^{(k)}\right)^2\right)\left(\eta_g^2 \tau_{\max}^2+1\right)\left(\sigma_{\ell}^2+\sigma_g^2+G\right) .
\end{aligned}
\end{equation}

\end{footnotesize}

\begin{equation}
    \begin{aligned}
    \mathbb{E}\left[T_1\right] 
    &\leq-\frac{\eta_g}{2}\left(\sum_{q=0}^{Q-1} \eta_{\ell}^{(q)}\right)\left\|\nabla f\left(w^{t'}\right)\right\|^2\\
    &+\sum_{q=0}^{Q-1} \frac{\eta_g \eta_{\ell}^{(q)}}{2} \mathbb{E}\left[T_3\right]\\
    &-\sum_{q=0}^{Q-1} \frac{\eta_g \eta_{\ell}^{(q)}}{2} \mathbb{E}_{\mathcal{H}}\left\|\frac{1}{m} \sum_{k=1}^m \nabla F_k\left(y_{k, q}^{t-\tau_k}\right)\right\|^2
\end{aligned}
\end{equation}
\begin{equation}
    \begin{aligned}
        \mathbb{E}\left[T_1\right] 
        &\leq-\frac{\eta_g \alpha(Q)}{2}\left\|\nabla f\left(w^{t'}\right)\right\|^2\\
        &+3 \eta_g L^2 Q \alpha(Q) \beta(Q)\left(\eta_g^2 \tau_{\max , K}^2+1\right)\left(\sigma_{\ell}^2+\sigma_g^2+G\right)\\
        &-\underbrace{\sum_{q=0}^{Q-1} \frac{\eta_g \eta_{\ell}^{(q)}}{2} \mathbb{E}_{\mathcal{H}}\left\|\frac{1}{m} \sum_{k=1}^m \nabla F_k\left(y_{k, q}^{t-\tau_k}\right)\right\|^2}_{T_4} .
    \end{aligned}
\end{equation}

Our method does not affect the value of T2 and its upper bound.
The T1 that would be affected in our approach was shown to have the same upper bound as Fedbuff.
\begin{footnotesize}

\begin{equation}
    \begin{aligned}
        \begin{aligned}
\mathbb{E}\left[T_2\right] 
& =\mathbb{E}\left[\frac{L \eta_g^2}{2 K^2}\left\|\sum_{k \in \mathcal{S}_t} \sum_{q=0}^{Q-1} \eta_{\ell}^{(q)} g_k\left(y_{k, q}^{t-\tau_k}\right)\right\|^2\right]  \\
&= \mathbb{E}\left[\frac{L \eta_g^2}{2 K^2}\left\|\sum_{k \in \mathcal{S}_t} \sum_{q=0}^{Q-1} \eta_{\ell}^{(q)}\left(g_k\left(y_{k, q}^{t-\tau_k}\right)-\nabla F_k\left(y_{k, q}^{t-\tau_k}\right)\right) \right.\right.\\
&\quad\left.\left.+ \sum_{k \in \mathcal{S}_t} \sum_{q=0}^{Q-1} \eta_{\ell}^{(q)} \nabla F_k\left(y_{k, q}^{t-\tau_k}\right)\right\|^2\right] \\
& \stackrel{\text { A.) })}{=} \frac{L \eta_g^2}{2 K^2} \mathbb{E}\left\|\sum_{k \in \mathcal{S}_t} \sum_{q=0}^{Q-1} \eta_{\ell}^{(q)}\left(g_k\left(y_{k, q}^{t-\tau_k}\right)-\nabla F_k\left(y_{k, q}^{t-\tau_k}\right)\right)\right\|^2\\
&+\frac{L \eta_g^2}{2 K^2} \mathbb{E}\left\|\sum_{k \in \mathcal{S}_t} \sum_{q=0}^{Q-1} \eta_{\ell}^{(q)} \nabla F_k\left(y_{k, q}^{t-\tau_k}\right)\right\|^2 \\
& \stackrel{\text { B.) }}{=} \frac{L \eta_g^2}{2} \sum_{k \in \mathcal{S}_t} \sum_{q=0}^{Q-1}\left(\eta_{\ell}^{(q)}\right)^2 \mathbb{E}\left\|\left(g_k\left(y_{k, q}^{t-\tau_k}\right)-\nabla F_k\left(y_{k, q}^{t-\tau_k}\right)\right)\right\|^2\\
&+\frac{L \eta_g^2}{2 K^2} \mathbb{E}\left\|\sum_{k \in \mathcal{S}_t} \sum_{q=0}^{Q-1} \eta_{\ell}^{(q)} \nabla F_k\left(y_{k, q}^{t-\tau_k}\right)\right\|^2 \\
& \leq \frac{L \eta_g^2 \beta(Q) \sigma_{\ell}^2}{2}\\
&+\frac{L Q \eta_g^2}{2 K} \sum_{k \in \mathcal{S}_t} \sum_{q=0}^{Q-1}\left(\eta_{\ell}^{(q)}\right)^2 \mathbb{E}_{\mathcal{H}} \mathbb{E}_{k \sim[m] \mid \mathcal{H}}\left\|\nabla F_k\left(y_{k, q}^{t-\tau_k}\right)\right\|^2 \\
& =\frac{L \eta_g^2 \beta(Q) \sigma_{\ell}^2}{2}\\
&+\frac{L Q \eta_g^2}{2 K} \sum_{k \in \mathcal{S}_t} \sum_{q=0}^{Q-1}\left(\eta_{\ell}^{(q)}\right)^2 \mathbb{E}_{\mathcal{H}}\left[\frac{1}{m} \sum_{i=1}^m\left\|\nabla F_i\left(y_{i, q}^{t-\tau_i}\right)\right\|^2\right] \\
& =\frac{L \eta_g^2 \beta(Q) \sigma_{\ell}^2}{2}\\
&+\underbrace{\frac{L Q \eta_g^2}{2 m} \sum_{q=0}^{Q-1} \sum_{i=1}^m\left(\eta_{\ell}^{(q)}\right)^2 \mathbb{E}_{\mathcal{H}}\left[\left\|\nabla F_i\left(y_{i, q}^{t-\tau_i}\right)\right\|^2\right]}_{T_5}
\end{aligned}
    \end{aligned}
\end{equation}

\end{footnotesize}

To ensure $T4 + T5 \leq 0$, it is sufficient to choose $\eta_g \eta_{\ell}^{(q)} Q \leq \frac{1}{L}$.
So,
% \clearpage
\begin{equation}
\begin{aligned}
     \mathbb{E}\left[f\left(w^{t+1}\right)\right] &\leq \mathbb{E}\left[f\left(w^{t'}\right)\right] \\
    &-\frac{\eta_g \alpha(Q)}{2}\left\|\nabla f\left(w^{t'}\right)\right\|^2\\
    &+3 \eta_g L^2 Q \alpha(Q) \beta(Q)\left(\eta_g^2 \tau_{\max}^2+1\right)\left(\sigma_{\ell}^2+\sigma_g^2+G\right)\\
    &+\frac{L}{2} \eta_g^2 \beta(Q) \sigma_{\ell}^2
\end{aligned}
\end{equation}

Summing up t from 1 to T and rearrange, yields,
\begin{equation}
    \begin{aligned}\sum_{t'=0}^{T-1} \eta_g \alpha(Q)\left\|\nabla f\left(w^{t'}\right)\right\|^2 & \leq \sum_{t=0}^{T-1} 2\left(\mathbb{E}\left[f\left(w^{t'}\right)\right]-\mathbb{E}\left[f\left(w^{t+1}\right)\right]\right)+\\&3 \sum_{t=0}^{T-1} \eta_g L^2 Q \alpha(Q) \beta(Q)\left(\eta_g^2 \tau_{\max}^2+1\right)\left(\sigma_{\ell}^2+\sigma_g^2+G\right) +\frac{L}{2} \eta_g^2 \beta(Q) \sigma_{\ell}^2 \\& 
    \leq 2\left(f\left(w^0\right)-f\left(w^*\right)\right) +3 \sum_{t=0}^{T-1} \eta_g L^2 \alpha(Q) \beta(Q)\left(\eta_g^2 \tau_{\max , K}^2+Q\right)\left(\sigma_{\ell}^2+\sigma_g^2+G\right)+\frac{L}{2} \eta_g^2 \beta(Q) \sigma_{\ell}^2
    \end{aligned}
\end{equation}

$\tau_{max} << T$, $0 \leq t-\tau_{\max} \leq t'\leq t \leq T$

\begin{equation}
    \begin{aligned}
        \begin{array}{r}\frac{1}{T} \sum_{t=0}^{T-1}\left\|\nabla f\left(w^{t}\right)\right\|^2 \leq \frac{2\left(f\left(w^0\right)-f^*\right)}{\eta_g \alpha(Q) T}+\frac{L}{2} \frac{\eta_g \beta(Q)}{\alpha(Q)} \sigma_{\ell}^2 +3 L^2 Q \beta(Q)\left(\eta_g^2 \tau_{\max}^2+1\right)\left(\sigma_{\ell}^2+\sigma_g^2+G\right) \end{array}
    \end{aligned}
\end{equation}

\end{document}